\documentclass[letterpaper, 10 pt, conference]{ieeeconf}  

\IEEEoverridecommandlockouts                              

\usepackage{graphicx}
\usepackage{amsmath} 
\usepackage{amssymb}  
\usepackage{makecell} 
\usepackage{booktabs} 
\usepackage{subcaption}

\title{\LARGE \bf
QuadBEV: An Efficient Quadruple-Task Perception Framework via Birds'-Eye-View Representation
}

\author{
    Yuxin Li$^{1,2}$ 
    Yiheng Li$^{1}$ 
    Xulei Yang$^{3}$ 
    Mengying Yu$^{2}$
    Zihang Huang$^{2}$ 
    Xiaojun Wu$^{2}$ 
    Chaikiat Yeo$^{1}$ 
    \thanks{*This work is sponsored by Desay SV Singapore}
    \thanks{1 School of Computer Science and Engineering, Nanyang Technological University, Singapore
        {\tt\small  yuxin004@e.ntu.edu.sg;}}%
    \thanks{2 Desay SV Automotive Singapore}%
    \thanks{3 Institute for Infocomm Research (I2R), Agency for Science, Technology and Research (A*STAR), Singapore}%
}
\begin{document}

\maketitle
\thispagestyle{empty}
\pagestyle{empty}


\begin{abstract}
Birds'-Eye-View (BEV) perception has become a vital component of autonomous driving systems due to its ability to integrate multiple sensor inputs into a unified representation, enhancing performance in various downstream tasks. However, the computational demands of BEV models pose challenges for real-world deployment in vehicles with limited resources. To address these limitations, we propose QuadBEV, an efficient multitask perception framework that leverages the shared spatial and contextual information across four key tasks: 3D object detection, lane detection, map segmentation, and occupancy prediction. QuadBEV not only streamlines the integration of these tasks using a shared backbone and task-specific heads but also addresses common multitask learning challenges such as learning rate sensitivity and conflicting task objectives. Our framework reduces redundant computations, thereby enhancing system efficiency, making it particularly suited for embedded systems. We present comprehensive experiments that validate the effectiveness and robustness of QuadBEV, demonstrating its suitability for real-world applications.
\end{abstract}
\section{Introduction}
Bird's-eye-view (BEV) perception is increasingly recognized as an essential technology within autonomous driving systems. By fusing data from multiple sensors, BEV techniques \cite{li2022bevformer, 2020liftsplatshoot, liu2022petr} provide a comprehensive, top-down view representation, which enhanced environmental perception across various tasks \cite{lanenet2018davy, maptr2023liao, flashocc2023yu, huang2021bevdet}. However, the computational intensity of traditional BEV methods often limits their deployment in vehicles with restricted computational resources, underscoring the need for efficient BEV-based perception frameworks that can handle multiple tasks without compromising performance.

Recent research has underscored the potential of multitask learning in autonomous driving. While a few methods \cite{liu2022bevfusion, liu2022petrv2} attempted to integrate multiple tasks into a unified framework, these efforts often prioritize model complexity over efficiency and have been limited to combining only two tasks. Tasks such as 3D object detection, lane detection, occupancy prediction, and map segmentation share substantial spatial and contextual information, suggesting significant untapped potential within multitask learning. A unified framework could lead to several benefits:

\textit{Mutual Information Exchange}: Features learned for one task can potentially enhance the performance of others. For instance, accurately detected lanes can improve the precision of object localization, while object detection can aid in identifying lane boundaries or occupancy regions.

\textit{Computational Efficiency}: By employing shared feature representations, a multitask framework can decrease redundant computations, thereby enhancing system efficiency and suitability for embedded systems.

Despite these potential benefits, progress in multitask perception frameworks remains limited. The integration of multiple perception tasks within a single framework introduces significant challenges, notably:

\textit{Learning Rate Sensitivity}: Different tasks may respond variably to the same learning rates, where an optimal rate for one could impede another's performance.

\textit{Conflicting Task Objectives}: Each task may require emphasis on different feature aspects, potentially leading to conflicts during training. For example, precise localization needed for object detection may conflict with the broader spatial understanding required for map segmentation or occupancy prediction.

In response to these challenges, we propose QuadBEV, an efficient, quadruple-task perception framework utilizing BEV representation. QuadBEV innovatively integrates four critical tasks: 3D object detection, lane detection, map segmentation, and occupancy prediction. This framework advocates for a streamlined structure that combines task-specific heads with a shared backbone, effectively utilizing BEV capabilities while simplifying the integration process through strategic multi-source dataset alignment and a tailored training regimen designed to manage task sequence and loss conflicts.

Our key contributions can be summarized as follows:
\begin{enumerate}
\item \textbf{Multitask Architecture}: We introduce a pioneering framework that comprehensively addresses four fundamental tasks in autonomous driving within a BEV framework: 3D object detection, lane detection, map segmentation, and occupancy prediction.
\item \textbf{Progressive Training Strategy}: We employ a carefully crafted training strategy with staged learning rate adjustments and a gradient-based loss balancing technique to facilitate balanced learning across diverse tasks.
\item \textbf{Experimental Validation}: Extensive testing validates the efficacy and robustness of QuadBEV, confirming its potential applicability in real-world autonomous driving scenarios.
\end{enumerate}

\begin{figure*}[ht]
\begin{center}
\includegraphics[scale=0.85, width=0.85\linewidth]{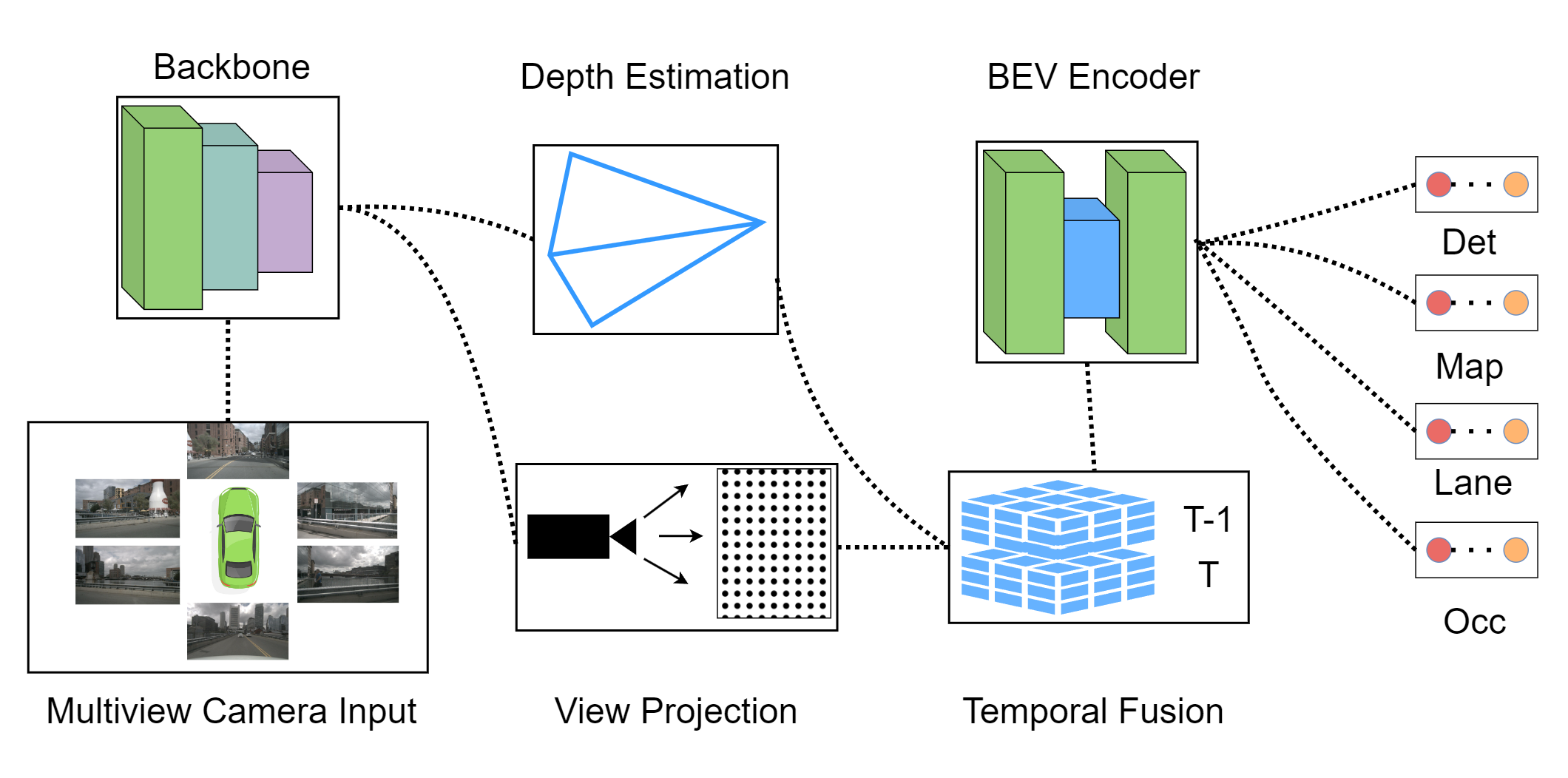}
\end{center}
\caption{Overall Architecture. Components in this architecture can be divided into two groups, shared feature extractors and task-specific heads. Shared feature extractors include 5 modules, backbone, depth estimator, view projector, temporal fusor and BEV encoder. Task-specific heads include 3d object detection, map segmentation, lane detection and occupancy prediction}
\label{fig:training_arch}
\end{figure*}

\section{Related Work}
\label{sec:related_work}
This section discusses key advancements and challenges in BEV perception and multi-task learning, essential for progress in autonomous driving technologies.

\textbf{Bird's-Eye-View Perception}
The BEV approach has gained considerable traction in sensor fusion applications, offering a novel methodology by integrating multiple sensor inputs within a pseudo-3D BEV framework. This innovative fusion technique transforms the traditional sensor alignment challenge into a manageable data-centric learning task, significantly enhancing the efficacy of various perception processes. Herein, we describe seminal contributions across distinct perception tasks within this domain.

\textbf{3D Detection in BEV}
3D object detection, crucial for autonomous navigation, has evolved significantly with the introduction of pioneer BEV models \cite{li2022bevformer, huang2022bevdet4d}, which integrate temporal dynamics and multi-view data to improve detection accuracy. Successive models such as \cite{li2022bevdepth, zong2023hop} have focused on leveraging depth cues and anticipatory frame analysis to refine detection capabilities. Later developments like \cite{lin2022sparse4d, cape2023xiong} have aimed to improve object interaction understanding in sparse environments, bringing vision-based systems on par with LiDAR-based alternatives.

\textbf{Segmentation in BEV}
BEV map segmentation has also advanced with approaches like HDMapNet \cite{2021hdmapnet}, introducing vectorized element prediction, and BEVSegFormer \cite{2022bevsegformer}, utilizing transformer architecture for distortion-resistant real-time segmentation. Further innovations by VectorMapNet \cite{li2022vectormapnet} and MapTR \cite{maptr2023liao} have enhanced element alignment and inference speed, pushing the boundaries toward effective mapless navigation.

\begin{figure*}[ht]
\begin{center}
\includegraphics[scale=0.85, width=0.85\linewidth]{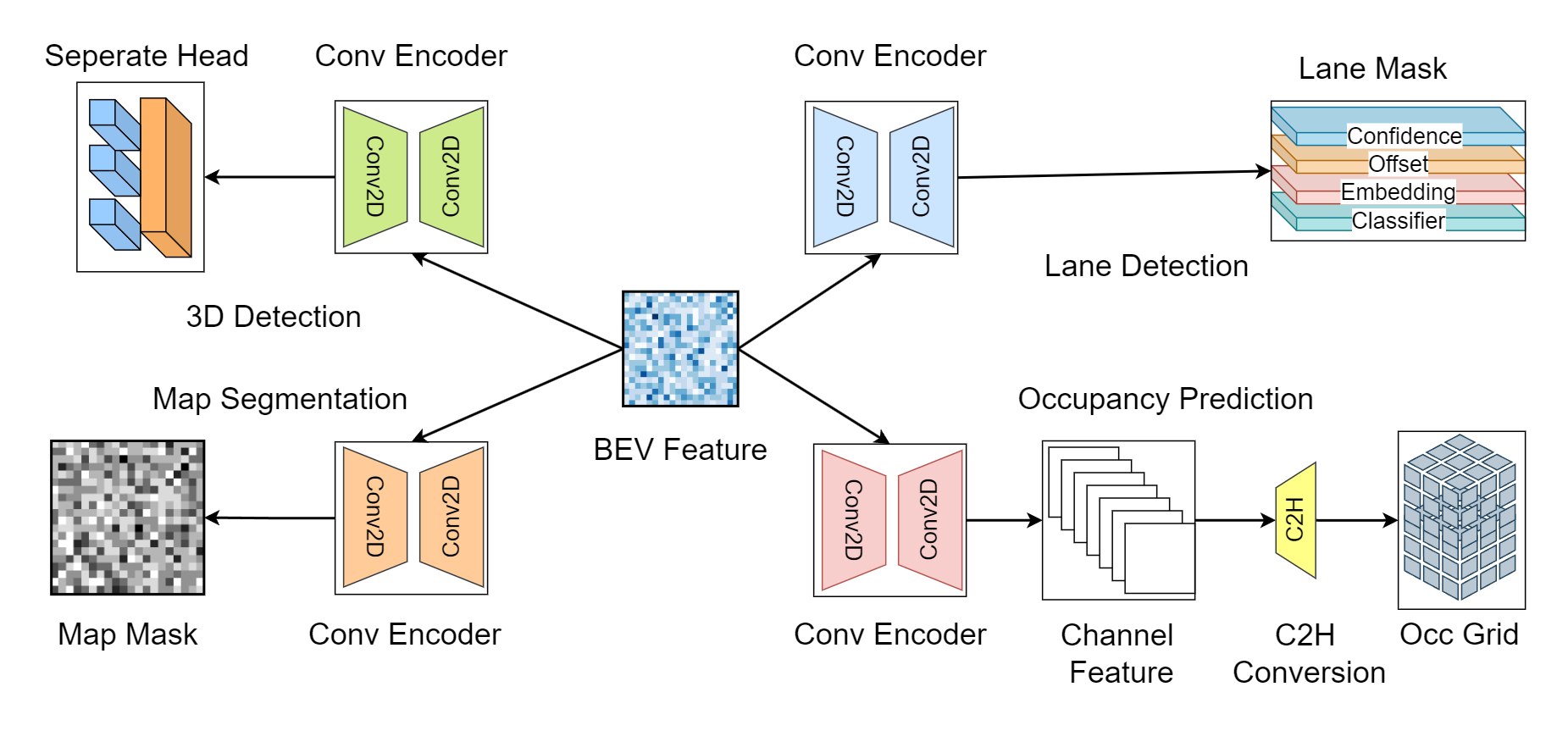}
\end{center}
\caption{Architecture of Quadruple Head on Shared BEV Feature. Four independent heads are attached to the BEV feature map in a round-robin manner.}
\label{fig:head_arch}
\end{figure*}

\textbf{Lane Detection in BEV}
Lane detection has transitioned from 2D to BEV-based methods, with developments such as \cite{bevlanedet23, anchor3dlane23} enhancing lane line prediction in 3D coordinates to address surface irregularities, signifying a shift towards a three-dimensional understanding of road structures.

\textbf{Occupancy Prediction in BEV}
Occupancy prediction is increasingly critical for autonomous driving, particularly for identifying atypical long-tail objects. Recent approaches like \cite{monoscene2022cao, occformer2023zhang, tpvformer2023huang, flashocc2023yu} have adopted voxel-based 3D space discretization for robust occupancy prediction, complementing traditional perception methods and offering reliable alternatives in complex scenarios.

\textbf{Multi-Task Learning}
Multi-task learning, a longstanding topic of interest, strives to mitigate the complexities of designing network architectures capable of handling concurrent tasks. Within autonomous perception, multi-task learning is strategically utilized to reduce computational demands and minimize task interference, adhering to the strict resource constraints of such systems. The spectrum of methods range from hard and soft parameter \cite{mtl_1, gradnorm2018chen} sharing to cross-stitch networks \cite{misra2016crossstitch}, task grouping \cite{taskgrouping_1, taskgrouping_2}, routing methodologies, and unified end-to-end training techniques \cite{e2e_mtl}. In the autonomous perception landscape, there is a call for more sophisticated integration. Preliminary efforts have attempted to unify tasks like map segmentation and object detection, yet a comprehensive integration within the BEV paradigm remains a promising research frontier.

\section{Methodology}
\label{sec:methodology}
Our methodology centres around a multi-task learning framework tailored for the diverse perception tasks in the BEV space. The architecture, training method, loss design, and evaluation metrics are carefully crafted to address the challenges and needs of autonomous driving.

\subsection{Model Architecture}
As illustrated in Figure \ref{fig:training_arch}, our architecture comprises six distinct modules, which are categorized into two primary groups: Shared Feature Extractors and Task-Specific Heads.

\textbf{Shared Feature Extractors}
These extractors are pivotal in synthesizing dense BEV semantic feature maps from multiview camera inputs. Initially, a convolutional-only backbone model \cite{wang2022yolov7} is applied to process the inputs iteratively. This is followed by the application of a view converter combining \cite{2020liftsplatshoot, huang2021bevdet} to transition from 2D to a pseudo-3D domain. Parallel to that, a fully-convolutional depth estimator is applied to refine the depth value in 3D space. Subsequent temporal information fusion is achieved via a concatenation module \cite{huang2022bevdet4d}, which is then further processed by another convolutional encoder \cite{liu2022bevfusion} to refine and finalize the BEV feature map.

\textbf{Task-Specific Heads}
\textit{Overall Architecture}: The architecture of our model comprises four distinct task-specific heads, each connected to a shared BEV feature map. As depicted in Fig. \ref{fig:head_arch}, these heads are arranged in a round-robin configuration without specialized routing architectures for different tasks. 

\textit{Convolutional Encoders} Each of the four task-specific heads is preceded by an identical yet independent convolutional encoder, tailored to refine the BEV features into more nuanced, task-sensitive features. These encoders are intentionally lightweight and consist of three convolutional layers, with slight variations in the number of input and output channels to meet the specific resolution requirements of each downstream task.

\textit{3D Object Detection:} This head aims to determine the center, scale, and orientation of objects within the scene. The design follows the configuration of CenterPoint \cite{centerpoint2021}, using a stack of convolutional modules to predict object parameters such as position (x, y, z), dimensions (w, h, l), and yaw angle.

\textit{Map Segmentation:} The map segmentation head is tasked with creating masks for various road elements, framing the problem as a semantic segmentation task. This head outputs multiple binary masks, one for each class, using three convolutional layers, similarly structured to the segmentation branch of BEVFusion \cite{liu2022bevfusion}.

\textit{Lane Detection:} Focused on identifying lane boundaries, this head utilizes four layers, both deconvolutional and convolutional, to process information related to the confidence, offset, embedding, and class of lane markers, mirroring the architecture used in BEVLaneDet \cite{bevlanedet23}.

\textit{Occupancy Prediction:} Addressing scenarios where 3D object detection might fail, such as with out-of-vocabulary or irregularly shaped objects, this head predicts the occupancy status of voxels in 3D space. It comprises a final convolutional layer followed by two MLP layers, configured as per the specifications in FlashOcc \cite{flashocc2023yu}.

\subsection{Training Method}
Our training methodology is structured into three distinct phases, each designed to progressively refine the model's capabilities across multiple tasks:

\textbf{Feature Extractor Pretraining}: Initially, we focus on developing a robust feature extractor by training an end-to-end model solely with the map segmentation task head using the NuImages dataset \cite{NuScenes2020} for M epochs. This phase is designed to cultivate a robust feature extractor that benefits all subsequent tasks by establishing a strong foundational representation.

\textbf{Sequential Training with Multitask Warm-Up}: Subsequently, we stabilize the feature extraction layers, specifically, the backbone, depth estimation, and BEV encoder modules, by freezing their parameters. Training then proceeds with all four task-specific heads attached, although they are categorized into primary and auxiliary roles. The primary task head is trained at a standard learning rate, while the auxiliary heads receive a reduced rate (one-tenth of the primary). This setup rotates among the four tasks to ensure each one serves as the primary task during different training intervals. This phase is designed to fine-tune the convolutional encoders within each head, allowing them to adapt to a multitask environment. Each training cycle in this phase lasts for N epochs.

\textbf{End-to-End Training with Gradient Weighting}: In the final phase, we eliminate the distinction between primary and auxiliary tasks, standardizing the learning rate across all tasks. Additionally, a separate, lower learning rate (one-tenth of the main rate) is applied to the backbone optimizer to promote stability. We initiate the training with equal loss weights for all tasks, which are then dynamically adjusted using a gradient-based weighting algorithm \cite{gradnorm2018chen}. This approach is intended to balance the training load among tasks, thereby minimizing loss variances and enhancing overall model performance.

\begin{table*}[t]
\caption{Comparison against SOTA 3D Detection Methods on NuScenes Val Dataset}
\label{table:sota_3d}
\begin{center}
\begin{tabular}{c|c|c|c|c| c c c c c}
\hline
Method & Input &  Latency(ms) & mAP $\uparrow$ & NDS $\uparrow$ & mATE $\downarrow$ & mASE $\downarrow$ & mAOE $\downarrow$ & mAVE $\downarrow$ & mAAE $\downarrow$ \\
\hline 
DETR3D \cite{detr3d}                 & 704x256 &    50.7 &    23.9 &    31.0 &    0.972 &    0.292 &    0.625  &    1.073 &    0.304 \\
PETR \cite{liu2022petr}              & 704x256 &    43.3 &    28.2 &    34.9 &    0.806 &    0.283 &    0.700  &    0.978 &    0.289 \\
BEVDet \cite{huang2021bevdet}        & 704x256 &    30.2 &    31.2 &    39.2 &    0.691 &    0.272 &    0.523  &    0.909 &    0.247 \\
CAPE \cite{cape2023xiong}            & 704x256 &    75.2 &    31.8 &    44.2 &    0.760 &    0.277 &    0.560  &    0.386 & \bf0.182 \\
BEVFormer \cite{li2022bevformer}     & 704x256 &    50.3 &    32.8 &    39.5 &    0.661 &    0.259 & \bf0.357  &    1.593 &    0.197 \\
BEVDet4D \cite{huang2022bevdet4d}    & 704x256 &    64.5 &    33.8 &    47.6 &    0.672 &    0.274 &    0.460  &    0.337 &    0.185 \\
TiGBEV \cite{tigbev2022huang}        & 704x256 &    68.0 &    35.6 &    47.7 &    0.648 &    0.273 &    0.517  &    0.364 &    0.210 \\
BEVDepth \cite{li2022bevdepth}       & 704x256 &    53.8 &    35.7 &    48.1 &    0.609 &    0.262 &    0.511  &    0.408 &    0.202 \\
HoP \cite{zong2023hop}               & 704x256 &    80.6 &    36.8 &    48.3 &    0.643 &    0.289 &    0.551  &    0.312 &    0.216 \\
SOLOFusion \cite{solofusion2022park} & 704x256 &    90.1 &    42.7 &    53.4 &    0.567 &    0.274 &    0.411  & \bf0.252 &    0.188 \\
\hline 
Ours(Baseline)                       & 704x256 & \bf21.0 & \bf45.6 & \bf55.5 & \bf0.549  &   0.278 &    0.438  &    0.270 &    0.196 \\
Ours(Quad)                           & 704x256 &    79.2 &    45.4 &    55.2 &    0.552  &   0.275 &    0.441  &    0.276 &    0.201 \\
\hline
\end{tabular}
\end{center}
\textbf{NOTE}: To ensure fair comparison, we re-produced results for all methods using open-source code at 704x256 resolution (model weights larger than this were absent from the original code release). This applies to all other tables unless otherwise noted. For the last two table rows, 'Baseline' refers to a baseline model trained independently, while 'Quad' indicates multiple tasks trained jointly. Latency for 'Quad' model is measured by running all four tasks together.
\end{table*}

\begin{table}[t]
\caption{Comparison with SOTA Map Segmentation Methods on NuScenes Val Dataset}
\label{table:sota_map}
\begin{center}
\setlength\tabcolsep{3pt} 
\begin{tabular}{c|c|c|c|c c c c c}
\hline
Method                                      & \rotatebox[origin=c]{90}{Input} 
                                            & \rotatebox[origin=c]{90}{Latency(ms)}
                                            & \rotatebox[origin=c]{90}{Mean IoU} 
                                            & \rotatebox[origin=c]{90}{Drivable} 
                                            & \rotatebox[origin=c]{90}{Road Marker} 
                                            & \rotatebox[origin=c]{90}{Walkway}   
                                            & \rotatebox[origin=c]{90}{Carpark}   
                                            & \rotatebox[origin=c]{90}{Divider} \\
\hline
Image2Map \cite{image2map2022}               & C     &    43.8      & 25.0       & 72.6      & 36.3      & 32.4      & 30.5      & -     \\
CVT \cite{cvt2022zhou}                       & C     &    703.6     & 40.2       & 74.3      & 36.8      & 39.9      & 35.0      & 29.4  \\
LSS \cite{2020liftsplatshoot}                & C     &    49.5      & 44.4       & 75.4      & 38.8      & 46.3      & 39.1      & 36.5  \\
BEVSegFormer \cite{2022bevsegformer}         & C     &    233.5     & 44.6       & 50.0      & 32.6      & -         & -         & 51.1  \\
Ego3RT \cite{ego3rt2022lu}                   & C     &    267.3     & 55.5       & 79.6      & 48.3      & 52.0      & 50.3      & 47.5  \\
BEVFusion \cite{liu2022bevfusion}            & C     & \bf23.2      & 56.1       & 81.2      & 54.2      & 57.6      & 50.7      & 45.9  \\
\hline
PointPillars \cite{pointpillar2018}          & L     &    102.9     & 43.8       & 72.0      & 43.1      & 53.1      & 27.7      & 37.5  \\
CenterPoint \cite{centerpoint2021}           & L     &    111.2     & 48.6       & 75.6      & 48.4      & 57.1      & 31.7      & 41.9  \\
\hline
HDMapNet \cite{2021hdmapnet}                 & C+L   &    194.2     & 44.5       & 56.0      & 31.4      & -         & -         & 46.1 \\
\hline
Ours(Baseline)                               & C     &    26.0      & 55.7       & 82.7      & 53.4      & 56.9   & \bf51.3      & 45.1 \\
Ours(Quad)                                   & C     &    79.2   & \bf56.4   & \bf 82.9   & \bf54.2  & \bf 57.5      & 51.2   & \bf46.4 \\
\hline 
\end{tabular}
\end{center}
\end{table}

\begin{table}[t]
\caption{Comparison with SOTA Lane Detection Methods on Openlanev2 Dataset}
\begin{center}
\setlength\tabcolsep{3pt} 
\begin{tabular}{l|c|c|c|c|c|c|c|c}
\hline
Method                              & \rotatebox[origin=c]{90}{Latency(ms)}            
                                    & \rotatebox[origin=c]{90}{F-Score}  
                                    & \rotatebox[origin=c]{90}{Up \& Down} 
                                    & \rotatebox[origin=c]{90}{Curve} 
                                    & \rotatebox[origin=c]{90}{Extreme Weather} 
                                    & \rotatebox[origin=c]{90}{Night} 
                                    & \rotatebox[origin=c]{90}{Intersection}
                                    & \rotatebox[origin=c]{90}{Merge \& Split} \\
\hline
3D-LaneNet \cite{3dlanenet2019noa} &    19.4    &    44.1 & - &    46.5  &    47.5     & 41.5   & 32.1    & 41.7           \\
GenLaneNet \cite{lanenet2018davy}  &    16.7    &    32.3 & - &    33.5  &    28.1     & 18.7   & 21.4    & 31.0           \\
PersFormer \cite{persformer23li}   &    33.5    &    50.5 & - &    55.6  &    48.6     & 46.6   & 40.0    & 50.7           \\
Anchor3DLane \cite{anchor3dlane23} &    19.5    &    54.3 & - &    58.0  &    52.7     & 48.7   & 45.8    & 51.7           \\
BEVLaneDet \cite{bevlanedet23}     & \bf14.3    &    58.1 & - & \bf63.2  & \bf53.4     & 53.1 &\bf50.3    & 53.4           \\
\hline
Ours(Baseline)                     &    17.8    &    57.8 & - &    62.9  &    52.9     & 53.2   & 48.8    & 52.9            \\
Ours(Quad)                         &    79.2    & \bf58.4 & - &    63.1  &    53.1  & \bf54.3   & 48.6 & \bf53.6            \\
\hline
\end{tabular}
\end{center}
\label{table:sota_lane}
\textbf{NOTE}: Since the OpenlaneV2-SubsetB does not contain height information, the performance of the Up \& Down section is omitted for fair comparison with other publications.
\end{table}

\begin{table*}[h]
\caption{Comparison with SOTA Occupancy Prediction Methods on Occ3D-NuScenes Validation Dataset}
\centering
\setlength\tabcolsep{4pt} 
\begin{tabular}{l|c|c| c c c c c c c c c c c c c c c c c}
\hline
Method      & \rotatebox[origin=c]{90}{Latency(ms)}
            & \rotatebox[origin=c]{90}{mIoU} 
            & \rotatebox[origin=c]{90}{others} 
            & \rotatebox[origin=c]{90}{barrier} 
            & \rotatebox[origin=c]{90}{bicycle} 
            & \rotatebox[origin=c]{90}{bus} 
            & \rotatebox[origin=c]{90}{car} 
            & \rotatebox[origin=c]{90}{Cons. Veh}
            & \rotatebox[origin=c]{90}{motorcycle} 
            & \rotatebox[origin=c]{90}{pedestrian} 
            & \rotatebox[origin=c]{90}{traffic cone} 
            & \rotatebox[origin=c]{90}{trailer} 
            & \rotatebox[origin=c]{90}{truck} 
            & \rotatebox[origin=c]{90}{Dri. Sur} 
            & \rotatebox[origin=c]{90}{other flat} 
            & \rotatebox[origin=c]{90}{sidewalk} 
            & \rotatebox[origin=c]{90}{terrain} 
            & \rotatebox[origin=c]{90}{manmade} 
            & \rotatebox[origin=c]{90}{vegetation}  \\               
\hline 
OccFormer~\cite{occformer2023zhang}&    193.3 &    20.3 &    5.5  &    27.8 &    11.2 &    31.9 &    34.0 &    12.8 &    15.7 &    16.5 &    8.0   &    12.3  &    23.4  &    47.8  &    28.4  &    31.8  &    20.2  &    6.4   &    6.1   \\  
TPVFormer~\cite{tpvformer2023huang}&    310.6 &    24.2 &    6.3  &    33.8 &    12.4 &    37.8 &    39.9 &    16.3 &    18.2 &    16.4 &    13.4  &    25.0  &    32.0  &    51.7  &    31.5  &    36.6  &    27.9  &    17.6  &    15.8  \\ 
CTF-Occ~\cite{occ3d2023tian}       &    156.0 &    24.8 &    7.4  &    34.2 &    17.8 &    35.5 &    37.6 &    14.7 &    21.8 &    19.7 &    18.3  &    20.8  &    27.0  &    49.6  &    30.8  &    33.0  &    28.9  &    18.4  &    16.4  \\ 
RenderOcc~\cite{renderocc2023pan}  &    242.5 &    25.2 &    4.2  &    27.6 &    9.3  &    24.0 &    22.9 &    12.6 &    15.8 &    15.3 &    15.4  &    19.6  &    20.2  &    58.7  &    31.7  &    42.0  &    43.2  &    16.9  &    18.0  \\   
PanoOcc~\cite{panoocc2023wang}     &    97.1  &    31.8 &    7.5  &    38.0 &    18.8 &    37.0 &    43.4 &    18.5 &    22.0 &    19.9 &    17.5  &    25.8  &    32.3  &    70.4  &    35.1  &    43.1  &    45.9  &    34.6  &    31.1  \\  
FlashOcc~\cite{flashocc2023yu}     & \bf47.2  & \bf37.8 & \bf11.3 & \bf44.1 & \bf23.9 & \bf43.5 &    48.5 &    24.1 & \bf25.3 &    24.9 & \bf25.9  &    32.7  & \bf37.8  &    74.3  & \bf40.6  & \bf49.1  & \bf50.8  &    44.5  & \bf38.9  \\ 
\hline
Ours(Baseline)                     &    56.4  &   36.3 &     11.0 &    42.8 &    23.6 &    42.7 &    48.0 & \bf24.4 &    24.5 &    24.7 &    24.9  &    31.8  &    36.7  &    74.7  &    39.5  &    47.5  &    48.3  &    44.0  &    38.3  \\
Ours(Quad)                         &    79.2  &   37.6 &     11.3 &    44.0 &    23.3 &    43.1 & \bf49.4 &    24.1 &    25.2 & \bf25.4 &    24.6  & \bf32.8  &    37.6  & \bf76.9  &    39.7  &    48.5  &    49.7  & \bf45.3  &    38.4  \\  
\hline
\end{tabular}
\label{table:sota_occ}
\end{table*}

\subsection{Loss Design}
Given the diversity and complexity of the tasks within our framework, an integrated and multifaceted loss function is crucial. The combined loss function, $L_{\text{combined}}$, encapsulates the contribution of each task-specific loss component, weighted appropriately to reflect their relative importance during training:

\[
L_{\text{combined}} = \alpha L_{3D} + \beta L_{\text{map}} + \gamma L_{\text{lane}} + \delta L_{\text{occ}} + \epsilon L_{\text{depth}}
\]

\textbf{Details of Loss Components}

\textit{3D Object Detection Loss} ($L_{3D}$): This component is crucial for the accurate localization and classification of 3D objects. It is defined as a combination of classification, regression, and IoU losses:
\[
L_{3D} = \lambda_1 L_{\text{cls}} + \lambda_2 L_{\text{reg}} + \lambda_3 L_{\text{iou}}
\]
where $L_{\text{cls}}$ employs binary cross-entropy (BCE) loss for object presence, $L_{\text{reg}}$ utilizes L1 loss for bounding box regression, and $L_{\text{iou}}$ incorporates IoU loss to enhance the overlap accuracy between predicted and actual object boundaries, as described in \cite{iouloss2019zhou}.

\textit{Map Segmentation Loss} ($L_{\text{map}}$): For the map segmentation task, focal loss \cite{focalloss2017lin} is used to refine the prediction of various map elements, focusing on reducing the imbalance between the more and less frequent classes.

\textit{Lane Prediction Loss} ($L_{\text{lane}}$): This loss facilitates the detection of lane markers and is articulated as:
\[
L_{\text{lane}} = \lambda_1 L_{\text{conf}} + \lambda_2 L_{\text{offset}} + \lambda_3 L_{\text{emb}} + \lambda_4 L_{\text{cls}}
\]
Here, $L_{\text{conf}}$ uses BCE loss for marker confidence, $L_{\text{offset}}$ applies mean squared error (MSE) loss for spatial deviations, $L_{\text{emb}}$ employs a PushPull loss to distinguish between different lane markers, and $L_{\text{cls}}$ leverages cross-entropy loss for classification tasks within lane detection \cite{bevlanedet23}.

\textit{Occupancy Prediction Loss} ($L_{\text{occ}}$): Cross-entropy loss is utilized to supervise the prediction of occupancy within a voxel grid, aiming to detect the presence of objects.

\textit{Depth Prediction Loss} ($L_{\text{depth}}$): This loss uses binary cross-entropy loss to guide the depth estimation process, which is vital for accurate 3D understanding from 2D inputs.

\textit{Dynamic Weighting}: The coefficients $\alpha, \beta, \gamma, \delta$, and $\epsilon$ balance the contributions of each loss component. Initially set to equal values, these weights are dynamically adjusted during training using the GradNorm algorithm \cite{gradnorm2018chen}, which helps maintain task balance and ensure uniform learning progress across all tasks.

\begin{table}[ht]
\caption{Comparison of Model Efficiency}
\begin{center}
\setlength\tabcolsep{6pt} 
\begin{tabular}{l|c|c|c c c c}
\hline
Method    & GFlops      & Latency  & Det   &  Map  & Lane  &  Occ\\ 
\hline
Fast-Sota & 578.2       & 158.5    &    45.6    &    56.1     &    58.1      & \bf37.8       \\
\hline
Baseline  & 645.7       & 169.7    & \bf45.6    &    55.7     &    57.8      &    36.3       \\
\hline                                    
Quad      & \bf281.3  & \bf79.2    &    45.4    & \bf56.4     & \bf58.4      &    37.6       \\
\hline
\end{tabular}
\end{center} 
Note: Fast-Sota methods are chosen based on their speed from among the top-performing models for each task, specifically, our baseline model for 3D det, BEVFusion for map segmentatin, BEVLaneDet for lane detection and FlasshOCC for occupancy prediction. Computational load (GFlops) and latency of both the Fast-Sota and Baseline models are measured by running four tasks concurrently on a single host machine equipped with an Nvidia 3090 GPU. The Quad model's performance metrics are evaluated in the same way using the QuadBEV framework.
\label{table:sota_latency}
\end{table}

\section{Experiments}
To demonstrate the effectiveness of our multitask architecture, we have conducted comprehensive experiments on public datasets and compared the results with other state-of-the-art (SOTA) methods. 

\subsection{Dataset}
To evaluate our methodology, we employed three publicly available datasets. NuScenes \cite{NuScenes2020}, containing data from a complete autonomous vehicle sensor suite (6 cameras, 5 radars, 1 lidar), was used for 3D object detection and map segmentation. NuScenes consists of 1000 scenes (20 seconds each), fully annotated with 3D bounding boxes (23 classes, 8 attributes). We utilized Occ3D \cite{occ3d2023tian} to benchmark occupancy prediction tasks. This large-scale dataset provides multi-view images and dense 3D voxel-based labels, annotating both occupancy status and semantic category. Occ3D builds upon popular autonomous driving datasets such as Waymo \cite{waymo2020} and NuScenes. OpenLaneV2 \cite{2023openlanev2wang}, designed for perception and reasoning in complex road scenes, was used for lane detection. In addition to lane information, OpenLaneV2 provides centerline and topology annotations for road elements. This dataset is also built from Waymo and NuScenes.

\subsection{Implementation Details}
For our research, we employed the MMDet3D framework \cite{mmdet3d2020} for all code implementations. We began by utilizing the camera branch from BEVFusion \cite{liu2022bevfusion}, replacing its backbone with ElanNet \cite{wang2022yolov7} to enhance feature extraction. Then we developed the depth estimator following BEVDepth \cite{li2022bevdepth} by replacing all modules in BEVDepth with convolutional neural networks. Additionally, we incorporated multi-scale BEV features using LSS-FPN as proposed by \cite{huang2021bevdet} and implemented a temporal fusion module based on BEVDet4D \cite{huang2022bevdet4d}. The BEV Encoder was adapted from the Resnet branch of BEVFusion.

For task-specific implementations, we developed the 3D object detection head using CenterPoint \cite{centerpoint2021}, which predicts 3D bounding boxes, and integrated CBGS \cite{cbgs2019} to enhance performance. The map segmentation task utilized the BEVSegHead from BEVFusion, while lane detection was carried out using BEVLaneDet \cite{bevlanedet23}, with modifications to its height prediction branch to classify each cell. The occupancy prediction head was implemented using FlashOcc \cite{flashocc2023yu}.

The datasets employed, NuScenes \cite{NuScenes2020}, Occ3D-NuScenes \cite{occ3d2023tian}, and OpenlaneV2-Subset-B (NuScenes) \cite{2023openlanev2wang}, share the common asset of the NuScenes dataset. We synchronized these datasets by timestamp matching to create a unified dataset supporting four tasks, and developed a wrapper to manage dataset fusion and training sample iteration. We initialized training with a depth prediction model using BEVDepth, generating and storing depth mask ground truths to optimize subsequent training phases. The input resolution was rescaled to $704 \times 256$, with data augmentation techniques including flipping, scaling, cropping, and rotation. Temporal fusion utilized BEV features from the previous two seconds, aggregating four frames at two frames per second.

Evaluation metrics included mAP and NDS (NuScenes Detection Score) for 3D detection, along with error metrics (mATE, mASE, mAOE, mAVE, mAAE) for various model performance aspects. Segmentation used the IoU metric, while lane detection and occupancy prediction employed the F1 score and mIoU, respectively.

Before initiating multi-task training, the backbone was first pre-trained on the NuImages dataset \cite{NuScenes2019}. We employed the AdamW optimizer throughout our training process. During the feature extractor training stage, we set a learning rate of 1e-4 and a weight decay of 1e-2, continuing for 20 epochs. Subsequently, for the multi-task warmup stage, the base learning rate was increased to 2e-4 with an auxiliary task-specific warmup learning rate of 2e-5; this stage lasted for 10 epochs per task. In the final end-to-end training stage, the learning rate for the base model was reverted to 1e-4 and the backbone fine-tuning rate was adjusted to 1e-5, with training extending over another 10 epochs. Training was conducted using a batch size of 8 on four A100 GPU cards. At each training stage, the best model weights were retained for fine-tuning in the subsequent phase.

\subsection{Task Specific Results}
Following a detailed description of our experimental setups, we present the outcomes in this section. It is pertinent to note that comparisons with state-of-the-art methods were conducted at a resolution of \(704\times256\), limited by the availability of larger model configurations and weights. Results were replicated using open-source codes.

In our experiments, the baseline model was configured to train on a single task utilizing our proposed feature extractor, while the quad mode involved training four task heads simultaneously in a multitask setting.

\textbf{3D Detection}
Our models exhibited exemplary performance in 3D object detection on the NuScenes Val Dataset, as detailed in Table \ref{table:sota_3d}. The baseline model attained an impressive mean Averag Precision (mAP) of 45.6\% and a NuScenes Detection Score (NDS) of 55.5\%, surpassing well-established methods such as SOLOFusion. The quad model also demonstrated robust performance, with only marginal reductions in mAP and NDS. This underscores our models' precision in detecting and localizing objects in three-dimensional space, a critical component of autonomous navigation safety and efficiency.

\textbf{Map Segmentation}
As shown in Table \ref{table:sota_map}, our map segmentation model performed robustly, particularly in complex urban settings on the NuScenes Val Dataset. The quad model achieved a superior Mean Intersection over Union (Mean IoU) of 56.4\%, slightly outperforming the baseline model. This efficacy highlights our model's ability to accurately identify and delineate diverse map features, which is vital for advanced driving assistance and autonomous operations.

\textbf{Lane Detection}
Our lane detection capabilities are detailed in Table \ref{table:sota_lane}, where the quad model excelled with the highest F-Score of 58.4\%. This indicates exceptional precision and recall in identifying lane markings under varied conditions, including adverse weather and nighttime scenarios, thereby ensuring the navigational safety of autonomous vehicles.

\textbf{Occupancy Prediction}
Table \ref{table:sota_occ} displays competitive performance in the occupancy prediction task on the Occ3D-NuScenes Validation Dataset. The quad model achieved an mIoU of 37.6\%, closely rivaling the leading FlashOcc model. This performance confirms our model's ability to effectively predict and delineate occupied spaces, providing a reliable foundation for obstacle detection and collision avoidance in autonomous driving systems.

In summary, our baseline model validates the efficacy of our foundational feature extractor and task-specific heads, performing comparably to or exceeding state of the art models. The outcomes from the quad model further demonstrate the success of our progressive training strategy, maintaining high-performance benchmarks with notable efficiency improvements.

\subsection{Latency Results}
Following the task-specific results, this section highlights the efficiency of our Quad model. Table \ref{table:sota_latency} presents a concise comparison of model efficiency. Remarkably, the Quad model reduces computational demands to just 281.3 GFlops while achieving a significantly reduced latency of 79.2 ms, substantially lower than both the Fast-Sota and Baseline models, which exhibit latencies of 158.5 ms and 169.7 ms, respectively. Despite these reductions, the Quad model sustains competitive performance across all tasks, underscoring its capability to deliver high-performance metrics efficiently, which is crucial for real-time processing in autonomous driving applications.

In conclusion, compared to conventional methods that train tasks individually before running them concurrently on a shared machine, our Quad model markedly enhances computational efficiency and processing speed while maintaining competitive benchmark performance against state of the art methods. 

\section{Abalation Study}

\begin{table}[ht]
\caption{Performance Variations against Pretraining Task }
\begin{center}
\setlength\tabcolsep{2pt} 
\begin{tabular}{l|c|c|c|c|c}
\hline
Method   &    Det-mAP &  Map-mIoU   & Lane-FScore  &    Occ-mIoU  &    Discount  \\ 
\hline
Baseline &    45.6    &    55.7     &    57.8      &    36.3      &    1.000   \\
\hline              
Det      &    44.3    &    54.8     &    55.2      &    36.5      &    0.917   \\
Map      & \bf45.4    & \bf56.4     & \bf58.4      &    37.6      & \bf1.055   \\
Lane     &    44.8    &    55.3     &    55.5      &    33.4      &    0.861   \\
Occ      &    45.2    &    47.7     &    49.3      & \bf37.9      &    0.756   \\
\hline
\end{tabular}
\end{center} 
NOTE: Column 'Discount' means the cumulative product of performance degradation when applying multitask training, compared to baseline model.
\label{tab:ab_order}
\end{table}

\begin{figure*}[h]
    \centering
    \begin{subfigure}{0.28\linewidth}
        \centering
        \includegraphics[width=\textwidth]{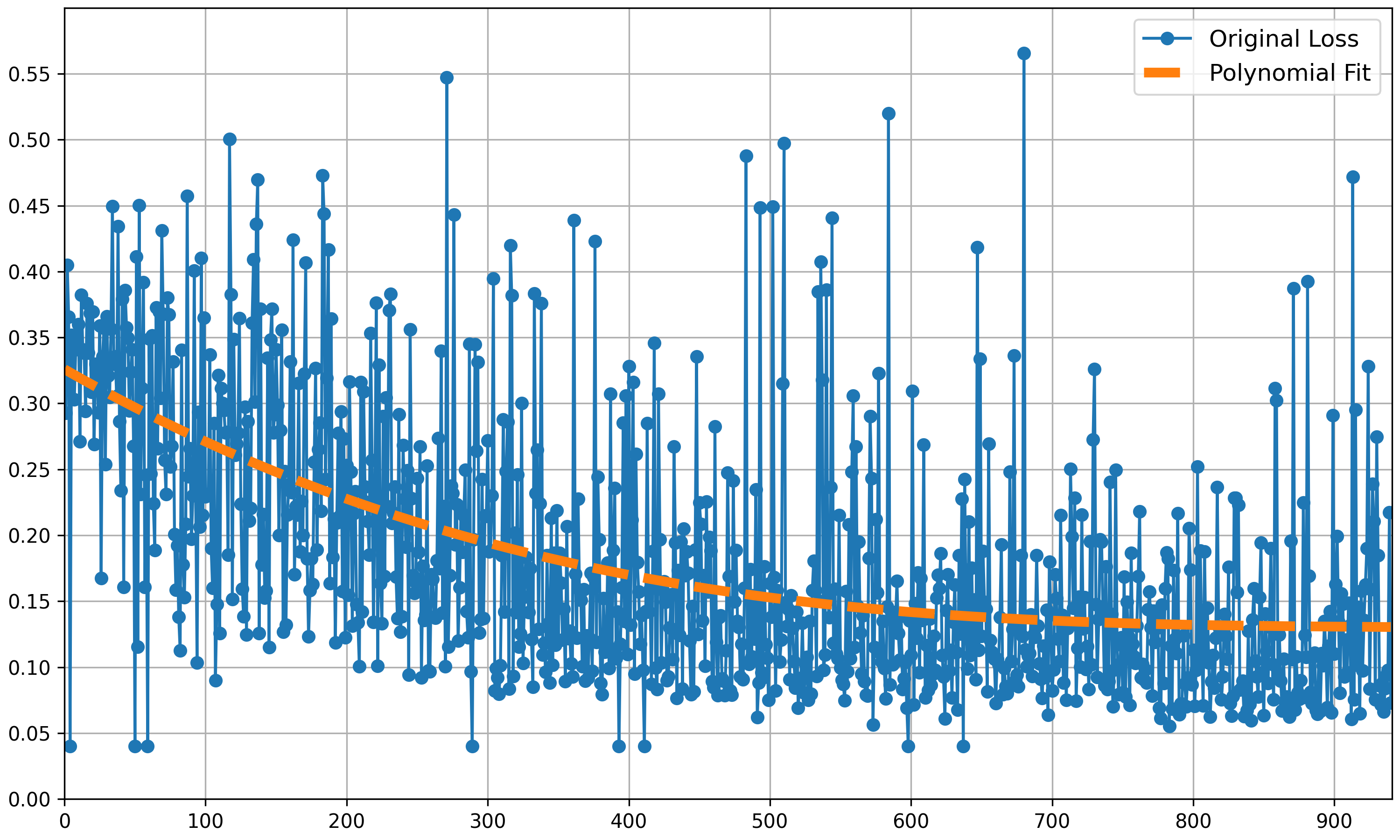}
        \caption{Baseline}
        \label{fig:ab_baseline}
    \end{subfigure}
    \hfill
    \begin{subfigure}{0.28\linewidth}
        \centering
        \includegraphics[width=\textwidth]{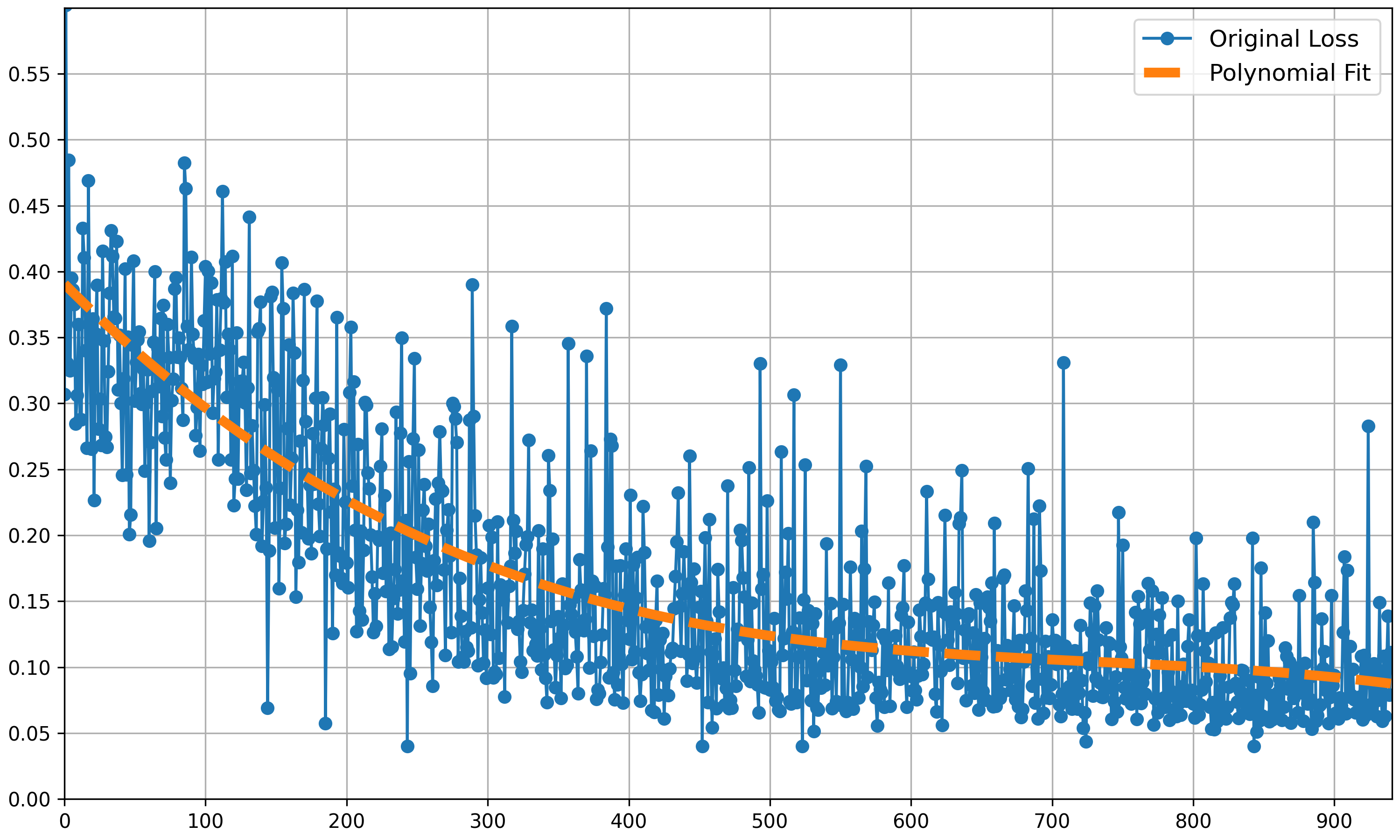}
        \caption{Pretraining with Map}
        \label{fig:ab_map}
    \end{subfigure}
    \hfill
    \begin{subfigure}{0.28\linewidth}
        \centering
        \includegraphics[width=\textwidth]{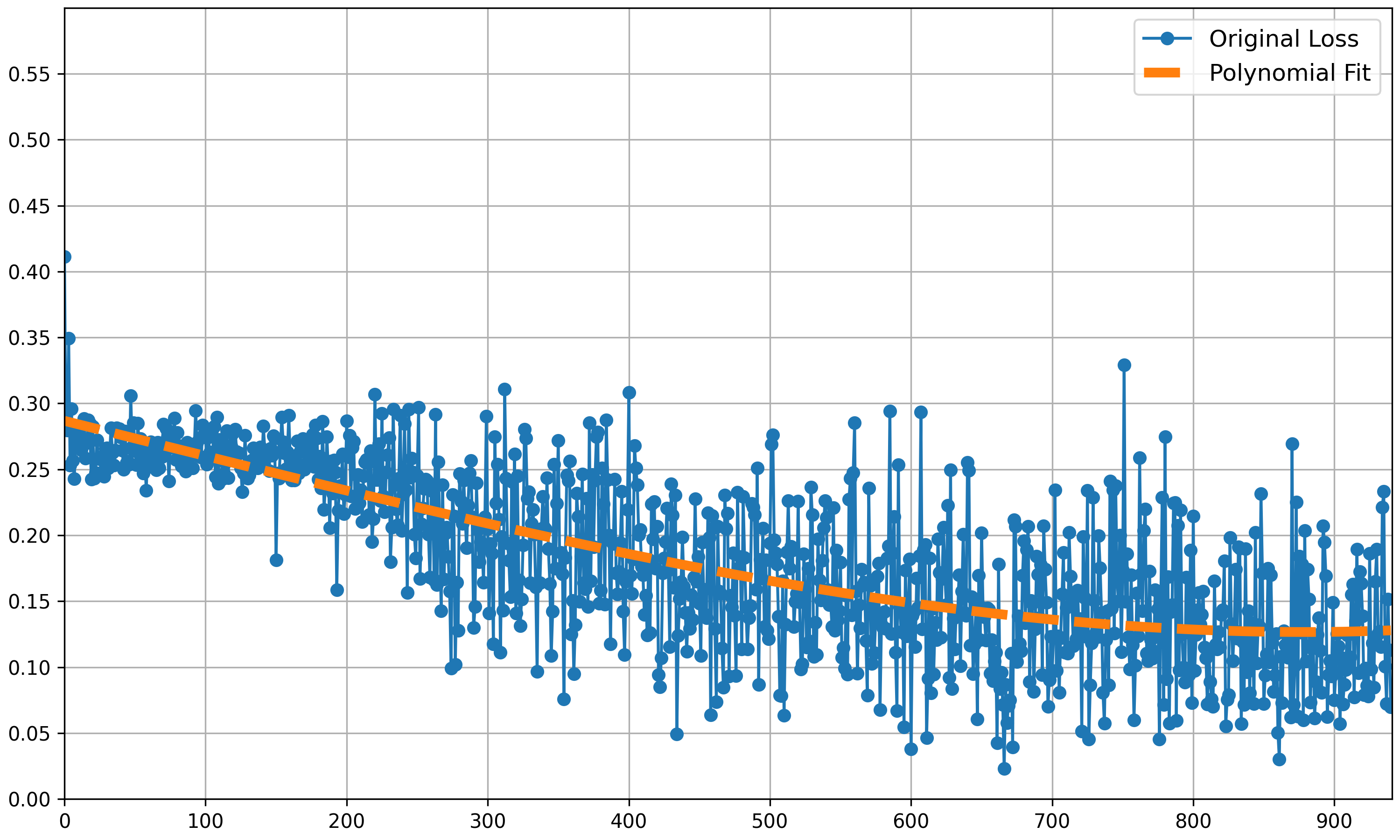}
        \caption{Multi-Head Warm Up}
        \label{fig:ab_warmup}
    \end{subfigure}

    \begin{subfigure}{0.28\linewidth}
        \centering
        \includegraphics[width=\textwidth]{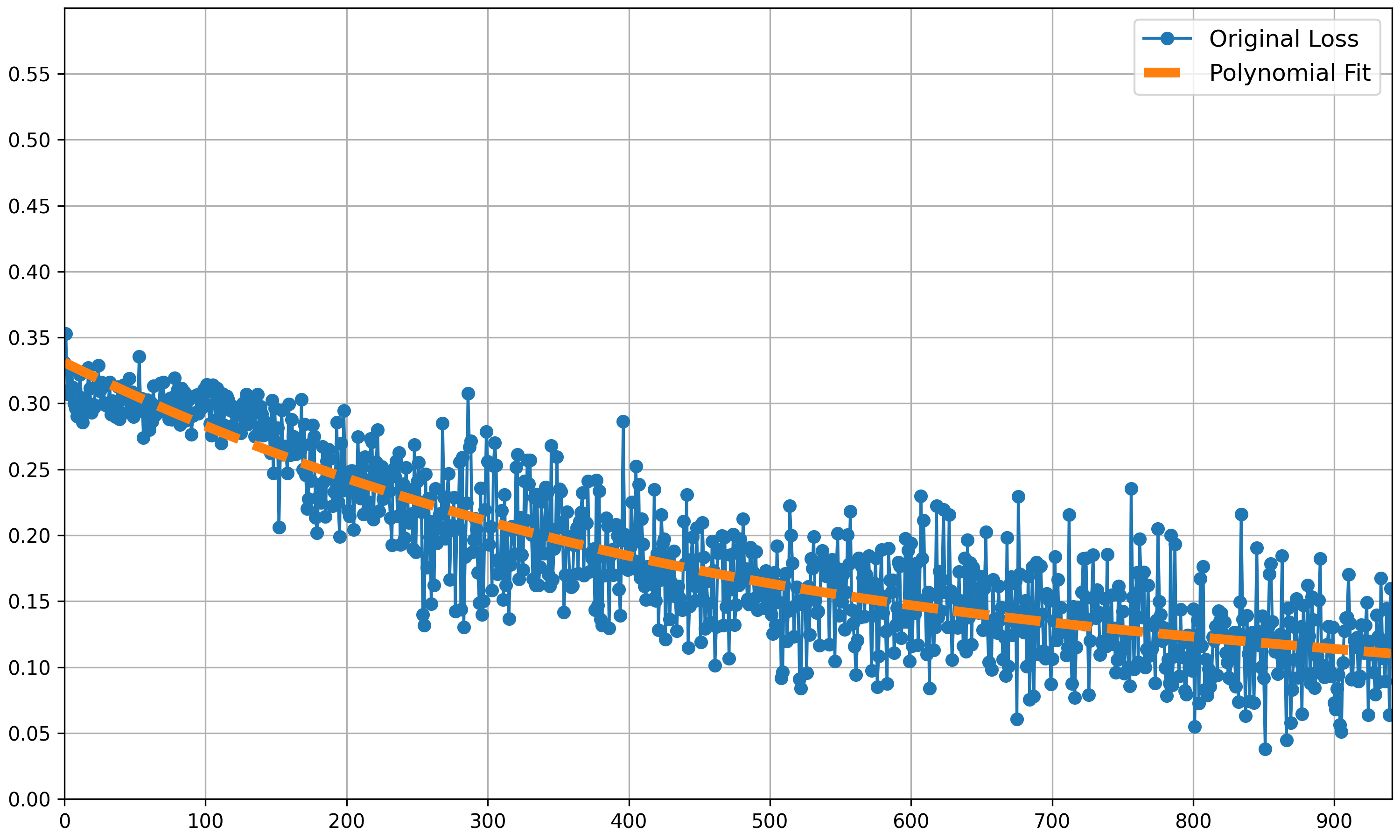}
        \caption{Backbone Enable Finetuning}
        \label{fig:ab_backbone}
    \end{subfigure}
    \hfill
    \begin{subfigure}{0.28\linewidth}
        \centering
        \includegraphics[width=\textwidth]{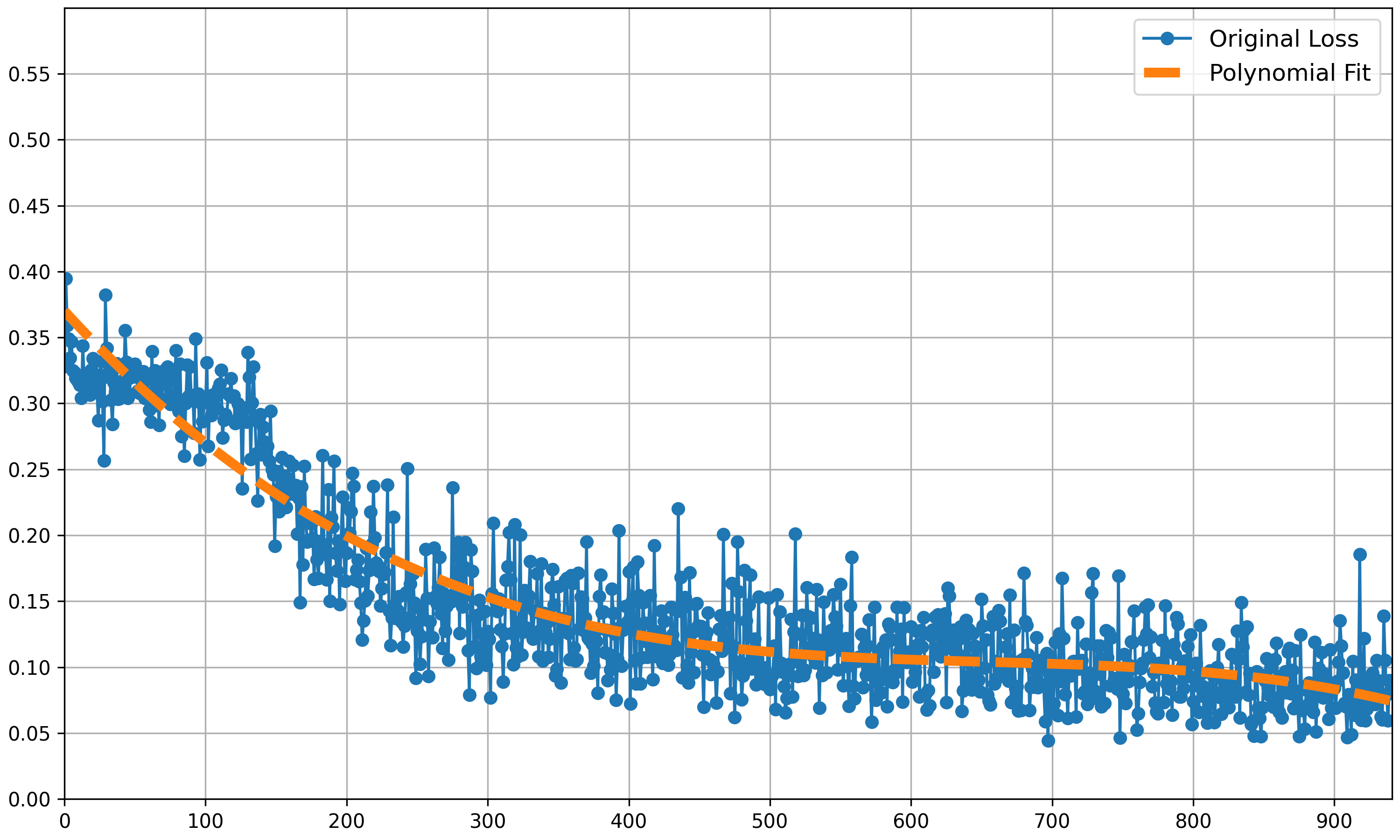}
        \caption{High Lane Weights}
        \label{fig:ab_weights}
    \end{subfigure}
    \hfill
    \begin{subfigure}{0.28\linewidth}
        \centering
        \includegraphics[width=\textwidth]{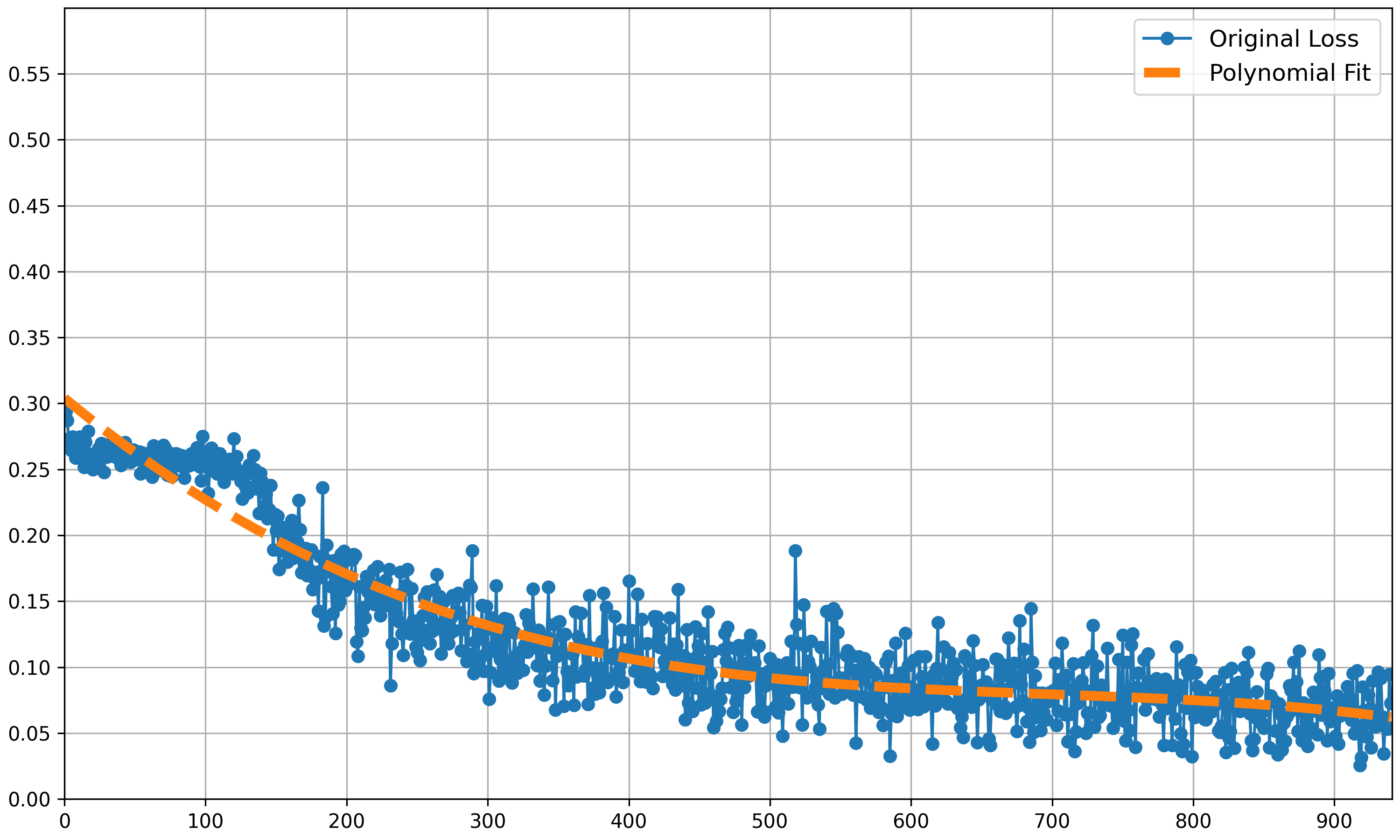}
        \caption{Gradient Weighting}
        \label{fig:ab_gradient}
    \end{subfigure}
    \caption{Comparisons between Different Learning Rate and Weights Schedule}
\end{figure*}

\subsection{Performance Variations against Pretraining Task}
This section examines the impact of varying the pretraining task within our multitask framework. We conducted an experiment where we alternated the pretraining task among four tasks, 3D Detection (Det), Map Segmentation (Map), Lane Detection (Lane), and Occupancy Prediction (Occ), and measured the average performance for each. Additionally, we calculated a "discount factor" to quantitatively assess the effects of different pretraining tasks on overall model performance. The discount factor is defined as the cumulative product of the performance degradation ratio when using a multitask model as compared to a baseline model.

Results, as summarized in Table \ref{tab:ab_order}, indicate that pretraining on the Map task resulted in the highest performance improvements in three out of the four tasks, as well as the highest discount factor. We hypothesize that this superior performance is due to the nature of the map segmentation task, which requires comprehensive pixel-level category prediction within the feature domain. This extensive feature interaction potentially facilitates a more balanced feature understanding, thus improving the model's adaptability to other tasks during subsequent fine-tuning phases.

\subsection{Comparison of Loss Profile against Progressive Training Strategy}
This section examines the impact of varying learning rate and weight schedules on the loss profiles of our models, thereby validating the efficacy of our progressive training strategy. Initially, the backbone model was trained uniformly across all tasks from scratch without any pretraining or warm-up phases.

In the baseline scenario, as depicted in Figure \ref{fig:ab_baseline}, we observed a highly unstable loss descent profile with significant variance in loss values. To mitigate this, we implemented backbone pretraining, which effectively reduced the loss variance as illustrated in Figure \ref{fig:ab_map}. Subsequently, during the second training stage, we initiated a warm-up period for the multitask head, which further diminished the loss variance, as shown in Figure \ref{fig:ab_warmup}. In our fourth experiment, fine-tuning of the backbone was conducted with a significantly lowered learning rate during the multitask training stage. Results, presented in Figure \ref{fig:ab_backbone}, showed a considerable reduction in loss variance and a record low in the minimum loss value compared to the baseline model.

Additionally, the fifth and sixth experiments explored the effects of manually assigning higher weights for lane detection and implementing the gradient-based weighting algorithm, GradNorm. The outcomes, displayed in Figures \ref{fig:ab_weights} and \ref{fig:ab_gradient}, revealed that the gradient-based weighting approach not only further reduced loss variance but also achieved the lowest minimum loss value. These experiments collectively demonstrate the robustness of our progressive training strategy and highlight the complexities involved in effectively training a multitask framework.

\section{Conclusion}
\label{sec:conclusion}
In this work, we introduced QuadBEV, an innovative and efficient multitask perception framework grounded in BEV representation. It unifies four fundamental autonomous driving tasks: 3D object detection, lane detection, map segmentation, and occupancy prediction. We demonstrate that a streamlined architecture and progressive training strategy are sufficient to manage the complexities of multitask learning. QuadBEV's ability to share representations and optimize training across tasks establishes it as a compelling solution for resource-constrained scenarios. Extensive evaluations underscore QuadBEV's robustness and potential for deployment in real-world autonomous driving systems. Future research directions could explore the integration of additional perception tasks and further refinement of training strategies for increasingly complex multitask scenarios.





\newpage

\bibliographystyle{IEEETran}
\bibliography{IEEEexample,myref}

\end{document}